\documentclass[twocolumn]{article}
\usepackage{PRIMEarxiv}

\usepackage[utf8]{inputenc} % allow utf-8 input
\usepackage[T1]{fontenc}    % use 8-bit T1 fonts
\usepackage{hyperref}       % hyperlinks
\usepackage{url}            % simple URL typesetting
\usepackage{booktabs}       % professional-quality tables
\usepackage{bm}
\usepackage{amsfonts}       % blackboard math symbols
\usepackage{amsthm}
\usepackage{amsmath}
\usepackage{amssymb}
\usepackage{mathtools}
\usepackage{microtype}      % micro typography
\usepackage{lipsum}
\usepackage{fancyhdr}       % header
\usepackage{graphicx}       % graphics
\graphicspath{{media/}}     % organize your images and other figures under media/ folder
\usepackage{comment}
\usepackage{here}
\usepackage{color}
\usepackage{multirow}
\usepackage{hhline}

\usepackage{tabularx}

\title{LARE: Latent Augmentation using Regional Embedding with Vision-Language Model}

\author{
  Kosuke Sakurai \\
  Waseda University \\
  \texttt{kosukesakurai@toki.waseda.jp} \\\
\And
  Tatsuya Ishii \\
  Waseda University \\
  \texttt{tishii2479@akane.waseda.jp} \\\
\And
  Ryotaro Shimizu \\
  Waseda University \\
  \texttt{shi3mizu8-r@fuji.waseda.jp} \\\
\And
  Linxin Song \\
  University of Southern California \\
  \texttt{linxinso@usc.edu} \\\
\And
  Masayuki Goto \\
  Waseda University \\
  \texttt{masagoto@waseda.jp} \\
}

\begin{document}
\twocolumn[
\maketitle

\vspace{-5mm}
\begin{abstract}
\vspace{3mm}
In recent years, considerable research has been conducted on vision-language models that handle both image and text data; these models are being applied to diverse downstream tasks, such as ``image-related chat,'' ``image recognition by instruction,'' and ``answering visual questions.''
Vision-language models (VLMs), such as Contrastive Language–Image Pre-training (CLIP), are also high-performance image classifiers that are being developed into domain adaptation methods that can utilize language information to extend into unseen domains.
However, because these VLMs embed images as a single point in a unified embedding space, there is room for improvement in the classification accuracy.
Therefore, in this study, we proposed the \textit{Latent Augmentation using Regional Embedding} (LARE), which embeds the image as a region in the unified embedding space learned by the VLM. 
By sampling the augmented image embeddings from within this latent region, LARE enables data augmentation to various unseen domains, not just to specific unseen domains. LARE achieves robust image classification for domains in and out using augmented image embeddings to fine-tune VLMs. 
We demonstrate that LARE outperforms previous fine-tuning models in terms of image classification accuracy on three benchmarks. We also demonstrate that LARE is a more robust and general model that is valid under multiple conditions, such as unseen domains, small amounts of data, and imbalanced data.
\end{abstract}

\vspace{3mm}
% keywords can be removed
\vspace{8mm}
]

%--------------------------------------------------------------------------------------------------------------------------------
%---------------------------------------------------------------------

\begin{figure*}[ht]
\centering
\includegraphics[width=0.9\linewidth]{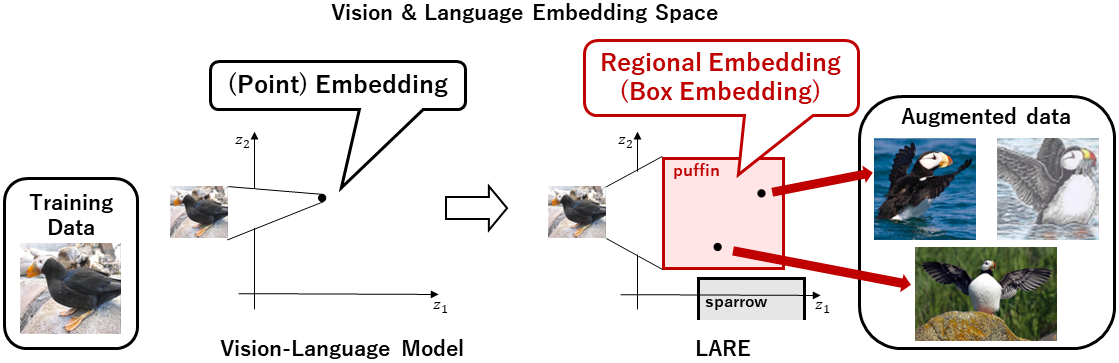}
\caption{Overview of \textit{Latent Augmentation using Regional Embedding} (LARE). LARE embeds the image as a region (box) in the vision-language embedding space instead of one embedding point like the classic vision-language model (VLM). A more robust image classification model can be constructed by fine-tuning the VLM, including the augmented embedding of various domains obtained from the latent region. Note that the augmented data on the right of the figure~\cite{CUB, CUB-Painting} is a hallucinated image and is not actually generated by LARE.}
\label{fig:1}
\end{figure*}

\section{Introduction}
Recent years, vision-language models (VLMs) such as Contrastive Language-Image Pre-training (CLIP)~\cite{CLIP}, Contrastive Captioner (CoCa)~\cite{CoCa}, and various other models~\cite{ALIGN, FILIP, Flamingo, BLIP, BLIP2} have shown outstanding generalizability on various downstream tasks.
Because VLMs have learned the relationship between texts and images, they are expected to perform well in image-classification task that leverage this knowledge.
In image classification task with a well-learned unified embedding space, users can easily and coarsely generalize these models by zero-shot classification, which directly calculates the similarity between label-image attention scores~\cite{CLIP, CoOp, Clip-adapter, Tip-adapter, UPL, Hazardous}.
However, such coarse-grain generalization cannot fully empower the model's performance in task-specific domains because of the model's text and class preferences and the lack of alignment between the text and image embedding space.
For example, CLIP performed well on images with patterns similar to those of other classes.

Therefore, a common practice for task-specific domains is fine-tuning, which trains a linear probe or multi-layer perception (MLP) aligned with the VLM's unified embedding space for downstream tasks~\cite{CLIP, Fine-tuning}.
To further enhance the performance of the task-specific linear probe or MLP, even in unseen domains that are not included in the training domain, the dataset can be augmented with synthetic images for unseen domains before fine-tuning~\cite{FAKE, ALIA, Attributed}. 
For example, by augmenting the image for unseen domains, such as ``painting,'' and ``snowy day'', the robust image classification model that can be adapted to augmented unseen domains is constructed.
Nonetheless, most image data augmentation methods rely on generative models, such as Stable Diffusion~\cite{StableDiffusion} or DALL-E~\cite{DALL-E, DALL-E2, DALL-E3}, and these models cannot faithfully follow a user's task-specific instructions. Therefore, they create unrelated and noisy images that adversely affect the performance of downstream tasks~\cite{liu2024best}.

In this study, we follow the strategy of utilizing language information from the unified embedding space learned by VLMs to augment the data of unseen domains in the latent space~\cite{LADS, TextManiA, Poda, LanDA}.
Augmenting data in the latent space not only generates data that follow a task-specific distribution but also leverages the semantic and domain knowledge of a unified vision-language embedding space.
For example, some studies can augment data (image embedding) in the direction of unseen domains in latent space by inputting text prompts, such as ``a painting of a [label],'' ``[label] on a snowy day'' into VLMs and can construct robust image classification models using augmented embedding to fine-tuning~\cite{LADS, TextManiA, LanDA}. 
However, these models can only augment data into one unseen domain per text prompt. In particular, they do not consider the diversity of the various domains in the test set because of overfitting to a specific domain.

Therefore, we propose \textit{Latent Augmentation using Regional Embedding} (LARE), a robust data augmentation method that applies regional embedding in the unified embedding space learned by the VLM.
In particular, as shown in Fig. \ref{fig:1}, LARE embeds the image as a region in the vision-language embedding space and augments data to various domains by sampling image embeddings from those regions. 
Data augmentation by regional embedding makes it possible to augment various unseen domains rather than just augmenting the specific unseen domain with only one text prompt ``a photo of a [label].''
To achieve regional embedding, we train a neural network that can transform each image embedding (a single point in the embedding space learned by the VLM) into a \textit{region} (\textit{box}~\cite{Word2Box, MBM, BoxTaxo}) in the latent space, which enlarges the size of the region while preserving the class-specific information of the original image. 
By fine-tuning the VLM with augmented data sampled from the region box, its performance in various unseen domains can be improved and a more robust and general model can be constructed.

We evaluated LARE using three benchmarks: CUB~\cite{CUB} (CUB-Painting~\cite{CUB-Painting}), DomainNet~\cite{DomainNet}, and CIFAR-100~\cite{CIFAR-100}. Our experimental results show that LARE outperforms previous fine-tuning models, such as CLIP, CoCa, and Latent Augmentation using Domain descriptionS (LADS)~\cite{LADS} in terms of image classification accuracy by up to 1.3\%. We also demonstrate that LARE outperforms previous models under multiple conditions, such as unseen domains, few-shot data, and imbalanced data. In addition, we compared the size and side lengths of the region (box) created by LARE and analyzed the usability of the region (box) for other tasks.
The main contributions are summarized as follows:
\begin{itemize}
    \item We proposed a novel image classification model, \textit{Latent Augmentation using Regional Embedding}, which can apply regional embedding (box embedding) to the VLMs. Using augmented data from the latent region, our method achieves a robust fine-tuning model that adapts to unseen domains.
    \item We introduce a novel domain adaptation method that can be augmented to various unseen domains without restrictions by leveraging the region with domain knowledge of VLM.
    \item We demonstrated that LARE outperforms previous methods under multiple conditions and identified the shape of the region, demonstrating that LARE is a more robust and general method.
\end{itemize}
%---------------------------------------------------------------------

%---------------------------------------------------------------------
\section{Related Work \label{sec:Related Work}}
\subsection{Vision-Language Model}
Vision-language models, such as CLIP~\cite{CLIP}, CoCa~\cite{CoCa}, and various other models~\cite{ALIGN, FILIP, Flamingo, BLIP, BLIP2}, are pre-trained models that embed images and languages in the same embedding space using large-scale image-language datasets. As it trains against language simultaneously, a unified embedding space can be used for various computer vision tasks.

\begin{figure}[t]
\centering
\includegraphics[width=1.0\linewidth]{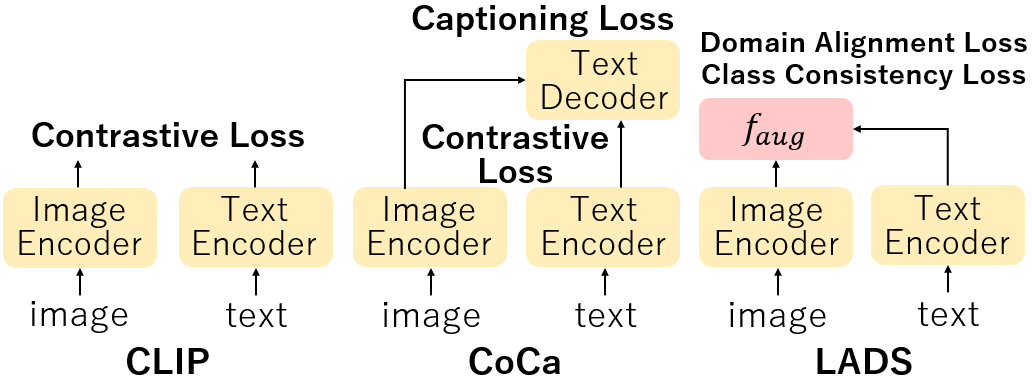}
\caption{Overview of CLIP, CoCa, and LADS}
\label{fig:2}
\end{figure}

CLIP is a multimodal model trained by contrastive learning~\cite{contrastive, Vilbert} using approximately 400 million pairs of images and captions such that the corresponding image and caption pairs are embedded at the same position in the embedding space. The CLIP structure is shown on the left in Fig. \ref{fig:2}. The CLIP utilizes a Transformer encoder~\cite{Transformer} as the text encoder and a Vision Transformer~\cite{Vit} as the image encoder. The property that similar image and text pairs are located in similar places in the embedding space makes it possible to perform zero-shot classification, where predictions are made using only prompts from class names without any additional training.

CoCa is a VLM that enables image classification and image captioning by adding the functions of SimVLM~\cite{SimVLM} to CLIP. By adding a caption generation function to CLIP, CoCa can consider the finer details of captions, resulting in a more accurate model than CLIP. The structure of CoCa is shown in the center of Fig. \ref{fig:2}. CoCa is trained using contrastive loss, such as CLIP, and captioning loss, such as SimVLM, which trains the output caption to be the same as the input caption and enables image captioning from image embedding. Consequently, a better embedding space can be utilized for downstream tasks.

%---------------------------------------------------------------------

%---------------------------------------------------------------------
\subsection{Domain Adaptation Method using Vision-Language Model}
Domain adaptation is the task of adapting models to perform well on unseen domains that are not included in the training data. Considerable research has been conducted on this topic~\cite{DA1, domain-adaptation1, domain-adaptation2, Visda, domain-adaptation3, ADDA, domain-adaptation4}. In the field of VLMs, considerable research has been conducted using pre-trained vision-language information for domain adaptation~\cite{AD-clip, DPL, WiSE-FT, Padclip, DAPrompt, ReCLIP, Promptstyler, Clipood}. In particular, augmenting image data in unseen domains for fine-tuning improves the image classification accuracy of such unseen domains while maintaining the fine-tuning accuracy of the training domain~\cite{FAKE, ALIA, Attributed, Styleclip, Stylegan-nada}. These methods can be used to augment data in unseen domains by inputting unseen text prompts into image generative models, such as Stable Diffusion~\cite{StableDiffusion} and DALL-E~\cite{DALL-E, DALL-E2, DALL-E3}.

However, collecting training data for every possible domain is expensive by directly generating images from scratch. This is because there is a cost to generate one image per text prompt as well as the cost of transforming the image to an image embedding through the VLM encoder. In particular, because there are countless domains to consider (e.g., differences in background or object numbers), the cost increases further with the number of unseen domains to be considered. Furthermore, it augments unrelated images and ignores task-specific information. Consequently, it is effective to utilize the unified embedding space learned by VLMs to augment the data of unseen domains in latent space~\cite{LADS, TextManiA, Poda, LanDA}. Data augmentation in the latent space can lower the training cost and allow the leveraging of the embedding space trained on large image-language data. For example, TextManiA~\cite{TextManiA}, LanDA~\cite{LanDA}, and LADS~\cite{LADS} obtained image embeddings of the unseen domain by shifting the image in the unseen domain direction, utilizing the unified embedding space with domain knowledge, and preserving task-specific information.

%---------------------------------------------------------------------

%---------------------------------------------------------------------
\subsection{Latent Augmentation using Domain descriptionS (LADS)}
LADS is an image classification model that extends CLIP to improve accuracy for specific unseen domains, which are difficult to obtain as training data, such as ``painting,'' and ``snowy day''. The structure of the LADS is shown on the right side of Fig. \ref{fig:2}. By inputting training image data and text prompts of training domains (e.g., ``a photo of a [label]'') and unseen domains (e.g., ``a painting of a [label],'' ``[label] on a snowy day'') into CLIP, LADS augments the image embedding of the unseen domain, which is a single point in the latent space. 

Specifically, LADS trains a neural network $f_{aug}$ (shown in Fig. \ref{fig:2}) that transforms the image embedding to a new image embedding of the unseen domain. A new image embedding is generated to transform the direction from the text prompt of the training domain to that of the unseen domain while preserving the original class information. By fine-tuning the CLIP, including the augmented data, it is possible to improve the performance of specific unseen domains while preserving the performance of the training domain. However, LADS can augment data to only one unseen domain per text prompt. Specifically, they do not consider diversity of various domains in the test set (e.g., differences in background or object numbers), owing to overfitting to a specific domain.
%---------------------------------------------------------------------

%---------------------------------------------------------------------
\section{Latent Augmentation using Regional Embedding (LARE)}

\begin{figure*}[ht]
\centering
\includegraphics[width=0.9\linewidth]{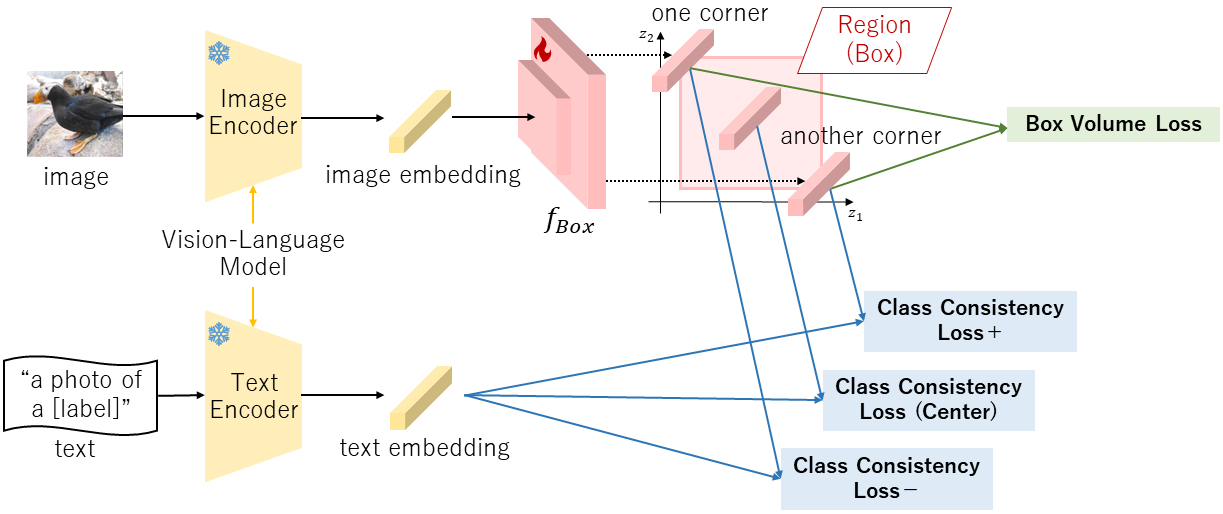}
\caption{Overview of Stage 1 in LARE. The network in Stage 1 outputs a region (box) in the latent space based on the embeddings obtained from the encoder of the VLM. The latent region is described by two points in the vision-language embedding space. The new neural network $f_{Box}$ is trained based on four losses: one box volume loss and three class consistency losses. After training the region, augmented data for unseen domains is created by randomly sampling from within the region (box).}
\label{fig:3}
\end{figure*}

In this study, we introduce \textit{Latent Augmentation using Regional Embedding} (LARE), a robust image classification model that applies regional embedding (box embedding~\cite{Word2Box, MBM, BoxTaxo}) in the unified embedding space trained by the VLM and augments data to various domains by sampling from those regions. Because LADS augments image embedding in the direction of the unseen domain while keeping it within the latent subspace of the original image classes, LARE represents the subspace of classes trained by the VLM as a region in the latent space. By sampling image embeddings from latent regions, it becomes possible to augment the data not only in specific unseen domains but also in various unseen domains without inputting specific text prompts of the unseen domains.

The LARE training process comprises two stages. In Stage 1, we train a neural network that transforms image embedding (a single point in the embedding space trained by VLM) to a region (box) in the latent space for each image. At this time, the region in the latent space is trained to (1) enlarge the region (box) while (2) retain the class information of the original image. In addition, we used CoCa as a VLM, contrary to LADS, to improve the accuracy. An overview of Stage 1 is presented in Fig. \ref{fig:3}.

In Stage 2, we fine-tune VLM on the training set containing both the original image embeddings and the augmented image embeddings to produce a classifier that is more robust to various domains. The augmented image embeddings are randomly sampled from within the region (box) generated in Stage 1, and the number of augmentations can be arbitrarily configured as a hyperparameter. Note that in both stages, we do not use any data from the unseen domain or text descriptions of the unseen domain as same as LADS, but only the text prompts of class names (e.g., ``a photo of [label]'').

\subsection{Stage 1: Learning the Region (Box)}
In Stage 1, we train a neural network $f_{Box}:\mathbb{R}^d\rightarrow\mathbb{R}^{2d}$, which transforms the image embedding into a region in the latent space for each image using the image and text embeddings obtained by VLM's image and text encoders. Here, $d$ is the dimension of the unified embedding space and a point representing an image in the embedding space embedded by VLM's image encoder is defined as

\begin{equation}
    \bm{x}=(x_1, x_2, \cdots, x_d)^T\in\mathbb{R}^d.
\end{equation}

We adopted box embedding~\cite{Word2Box, MBM, BoxTaxo} as the region in the latent space because of its simple structure, and the size and side lengths of the region can be easily calculated. 
For a representative element $\bm{x}$, the box is defined by the two corners of the box $\bm{X}^-=(X^-_1, X^-_2, \cdots, X^-_d)^T\in\mathbb{R}^d$ and $\bm{X}^+=(X^+_1, X^+_2, \cdots, X^+_d)^T\in\mathbb{R}^d$, and the region of a box $Box(\bm{x})$ is formulated as follows:

\begin{multline}
     {Box(\bm{x})} = \\
     \left\{(x_1, x_2, \cdots, x_d)^T \left| X^{-}_j  \le x_j \le X^{+}_j, \forall j \in \{ 1,2, \cdots, d \} \right\} \right.
\end{multline}

That is ${Box(\bm{x})} \subseteq \mathbb{R}^d$, and the network $f_{Box}:\mathbb{R}^d\rightarrow\mathbb{R}^{2d}$ from a represent element $\bm{x}$ to a box is equivalent to outputting the two corners of the box $\bm{X}^-=(X^-_1, X^-_2, \cdots, X^-_d)^T\in\mathbb{R}^d$, $\bm{X}^+=(X^+_1, X^+_2, \cdots, X^+_d)^T\in\mathbb{R}^d$ (two points in the embedding space).

In the training phase, the inputs in Stage 1 are image and text (``a photo of a [label],'' where label is defined as $y_i$ with $i=1,2,\cdots,n$), and the outputs are one corner $\bm{X}^-_i\in\mathbb{R}^d$ and another corner $\bm{X}^+_i\in\mathbb{R}^d$ of the region (box) for each image, where $n$ is the batch size.
First, images and texts are input to VLM's image and text encoders, respectively, to obtain image-embedded points $\bm{x}_i\in\mathbb{R}^d$ and text-embedded points  $\bm{T}_\theta(y_i)\in\mathbb{R}^d$, where $i = 1,2, \cdots, n$, and $\bm{T}_\theta(\cdot)$ is the output of the text encoder.
Second, the image-embedded point $\bm{x}_i$ is input to the additional network $f_{Box}$ and outputs the region (box) $\bm{X}^-_i$, $\bm{X}^+_i$. 
Finally, network $f_{Box}$ is trained using the region (box) $\bm{X}^-_i$, $\bm{X}^+_i$, and text embedding $\bm{T}_\theta(y_i)$. 
As aforementioned, a valuable box is (1) larger to include unseen domains while (2) preserving the class information of the original image. 
To achieve this, we trained $f_{Box}$ using a combination of two losses: {\it Box Volume Loss} and {\it Class Consistency Loss}.

\noindent
{\it Box Volume Loss}: Box volume loss encourages an increase in the box size. Generally, increasing the size of a box is equivalent to increasing its hypervolume~\cite{Word2Box}. However, because VLMs, such as CLIP and CoCa have embeddings located on a hypersphere through contrastive learning, we take a loss in reducing the cosine similarity of each corner of the box. Formally, box volume loss of $f_{Box}$ is defined as follows:

\begin{equation}
    L_{BV}(f_{Box})=\sum^n_{i=1}\left(\bm{X}^-_i\cdot\bm{X}^+_i\right),
    \label{eq:1}
\end{equation}

\noindent
where $\bm{A}\cdot\bm{B}$ is the inner product of embeddings $\bm{A}$ and $\bm{B}$. Note that each corner $\bm{X}^-_i$ and $\bm{X}^+_i$ in the vision-language embedding space was normalized to norm 1.

\noindent
{\it Class Consistency Loss}: Box volume loss generates boxes containing diverse unseen domains by increasing the box size. However, excessively large boxes lose the class information in the original image. Thus, we add class consistency loss, where each corner and center of the box preserve the class information. Each corner and center were trained to approximate the language embedding for a class in the original image, preserving class information across the entire region (box). Formally, class consistency loss of $f_{Box}$ is defined as

\begin{align}
    &L^-_{CC}(f_{Box})=\sum^n_{i=1}CE\left(S\left[\bm{X}^-_i\cdot{\bm{T}_\theta(y_i)}\right],y_i\right),
    \label{eq:2} \\
    &L^+_{CC}(f_{Box})=\sum^n_{i=1}CE\left(S\left[\bm{X}^+_i\cdot{\bm{T}_\theta(y_i)}\right],y_i\right),
    \label{eq:3} \\
    &L_{CC}(f_{Box})=\sum^n_{i=1}CE\left(S\left[\frac{\bm{X}^-_i+\bm{X}^+_i}{2}\cdot{\bm{T}_\theta(y_i)}\right],y_i\right),
    \label{eq:4}
\end{align}

\noindent
where $CE(a,b)$ is the cross-entropy loss between the predicted label $a$ and ground truth label $b$, and $S[\cdot]$ is the softmax function. Equation \eqref{eq:2} is trained for one corner $\bm{X}^-_i$, Equation \eqref{eq:3} for another corner $\bm{X}^+_i$, and Equation \eqref{eq:4} for the center of the box to maximize the similarity with the original class embedding via VLM zero-shot.

Our final objective function $L_{LARE}$ for train the neural network $f_{Box}$ in Stage 1 is a linear combination of box volume loss and class consistency loss:

\begin{multline}
    L_{LARE}(f_{Box})=(1-\alpha)L_{BV}(f_{Box}) \\
    +\alpha\left(\frac{L^-_{CC}(f_{Box})+L^+_{CC}(f_{Box})+L_{CC}(f_{Box})}{3}\right),
    \label{eq:5}
\end{multline}
\noindent

where $\alpha$ denotes a hyperparameter that determines the weight of each loss.

\subsection{Stage 2: Fine-tuning}

\begin{table*}[ht]
\begin{center}
\caption{In-domain, out-of-domain, and extended accuracy on CUB (CUB-Painting) and DomainNet. In-domain indicates accuracy on the same domain as the training set, out-of-domain indicates accuracy on unseen domains, and extended indicates accuracy on both domains. LARE (CoCa) outperforms all methods on CUB (CUB-Painting) and outperforms CoCa (fine-tuning) and LADS (CoCa) on DomainNet.}
\label{tab:1}
\scalebox{1.0}{
\begin{tabular*}{\linewidth}{@{\extracolsep{\fill}}c|rrr|rrr}
\toprule
& \multicolumn{3}{c|}{CUB (CUB-Painting)} & \multicolumn{3}{c}{DomainNet} \\ \midrule
Method & \multicolumn{1}{c}{In-domain} & \multicolumn{1}{c}{Out-of-domain} & \multicolumn{1}{c|}{Extended} & \multicolumn{1}{c}{In-domain} & \multicolumn{1}{c}{Out-of-domain} & \multicolumn{1}{c}{Extended}\\ \midrule
CLIP (zero-shot) & 63.27 & 55.10 & 60.40 & 93.38 & 96.09 & 95.62\\
CoCa (zero-shot) & 73.63 & 64.78 & 70.52 & 94.04 & \textbf{96.48} & \textbf{96.05}\\
CLIP (fine-tuning) & 86.42($\pm$0.05) & 65.31($\pm$0.09) & 78.85($\pm$0.06) & 96.74($\pm$0.04) & 92.68($\pm$0.03) & 93.33($\pm$0.02)\\
CoCa (fine-tuning) & 87.01($\pm$0.15) & 71.95($\pm$0.13) & 81.62($\pm$0.06) & 96.72($\pm$0.04) & 93.58($\pm$0.05) & 94.20($\pm$0.10)\\
LADS (CLIP) & 86.88($\pm$0.12) & 66.22($\pm$0.27) & 79.57($\pm$0.16) & 96.54($\pm$0.03) & 94.93($\pm$0.05) & 95.17($\pm$0.04)\\
LADS (CoCa) & 86.67($\pm$0.35) & 72.56($\pm$0.15) & 81.67($\pm$0.06) & 96.54($\pm$0.03) & 95.16($\pm$0.05) & 95.44($\pm$0.04)\\ \midrule
LARE (CLIP) & 87.01($\pm$0.10) & 65.99($\pm$0.30) & 79.63($\pm$0.03) & 96.58($\pm$0.13) & 95.00($\pm$0.06) & 95.27($\pm$0.03)\\
LARE (CoCa) & \textbf{87.03}($\pm$0.07) & \textbf{73.27}($\pm$0.41) & \textbf{81.94}($\pm$0.14) & \textbf{96.81}($\pm$0.10) & 96.11($\pm$0.03) & \textbf{96.05}($\pm$0.03)\\
\bottomrule
\end{tabular*}
}
\end{center}
\end{table*}

In Stage 2, we fine-tune the VLM on the training set containing both the original and augmented image embeddings, randomly sampled from the region (box) trained in Stage 1. We achieved this using linear probing as a fine-tuning technique, which trains only a linear classifier added to the final layer of the VLM image encoder. Using linear probing as a fine-tuning technique results in faster training and more robust classifiers~\cite{CLIP, Fine-tuning}. By performing linear probing, including augmented data from the region, we constructed a more robust image classification model that can adapt to various unseen domains.

%---------------------------------------------------------------------
\section{Experiment}
%---------------------------------------------------------------------
\subsection{Experimental Settings}

We conducted experiments using three datasets: CUB~\cite{CUB} (CUB-Painting~\cite{CUB-Painting}), DomainNet~\cite{DomainNet}, and CIFAR-100~\cite{CIFAR-100}. CUB and CUB-Painting are bird-image datasets containing 200 classes of real and painted images, respectively. We confirmed the accuracy of the unseen domain by predicting the data for CUB-Painting using the model trained on the CUB. Our DomainNet is a specific split~\cite{COAL} of the original DomainNet~\cite{DomainNet} dataset, which contains the 40 most common classes from four domains: ‘sketch,’ ‘real,’ ‘clipart,’ and ‘painting.’ Similar to prior work~\cite{LADS, Fine-tuning, COAL}, we train on ‘sketch’ and evaluate on the three other domains to confirm the unseen accuracy. CIFAR-100 is a dataset comprising color photographs of objects (such as plants, animals, equipment, and vehicles.) of 100 classes.

We compared LARE with three baselines: CLIP (zero-shot and fine-tuning), CoCa (zero-shot and fine-tuning), and LADS (CLIP and CoCa). The zero-shot in CLIP and CoCa uses only a text prompt (``a photo of a [label]'') to predict classes without training a model. Fine-tuning (linear probing) in CLIP and CoCa trains a linear classifier using only the original training data, without using augmented data. LADS (CLIP) and LADS (CoCa) use CLIP or CoCa as the backbone model and are fine-tuned by adding augmented data to a specific domain. For example, in the CUB-Painting dataset, LADS augments the training data for painting with the text prompt ``a painting of a [label].'' Note that LADS cannot be applied to the dataset CIFAR-100, which does not require shifting to a specific unseen domain, because it can only augment one or a few unseen domains.

We ran each method over five random seeds and reported the mean and standard deviation of the image classification accuracy. In our experiments, we employed AdamW~\cite{AdamW} with a batch size of 512 and the epoch was set to the maximum of the validation data. In LARE, the number of random samples from the region was set to 3 (CIFAR-100) or 5 (CUB) or 40 (DomainNet) depending on the size of the training dataset, training epoch of the neural network $f_{Box}$ to 100, and input text prompt to the text encoder to ``A photo of a [label].''

%---------------------------------------------------------------------

%---------------------------------------------------------------------
\subsection{Results}
%---------------------------------------------------------------------
\paragraph*{Result for Unseen Domain}
Table \ref{tab:1} lists the in-domain, out-of-domain, and extended accuracies of the CUB (CUB-Painting) and DomainNet. In-domain indicates accuracy in the same domain as the training set, out-of-domain indicates accuracy in unseen domains that are not included in the training domain, and extended indicates accuracy in both training and unseen domains.

The experimental results showed that LARE achieved the best accuracy for all domains in the CUB (CUB-Painting) dataset. In the DomainNet dataset, LARE (CoCa) outperformed CoCa (fine-tuning) and LADS (CoCa) in all domains, although it did not achieve CoCa (zero-shot) out-of-domain performance. LARE outperformed previous fine-tuning models in all domains, demonstrating that it is an effective data augmentation method. Furthermore, for the out-of-domain, LARE outperformed previous fine-tuning models by up to $2.5\%$. This suggests that LARE is an effective domain adaptation method for unseen domains.

\paragraph*{Results for CIFAR-100}
Table \ref{tab:2} shows the accuracy of CIFAR-100 compared with CoCa. The experimental results show that LARE outperforms CoCa (fine-tuning), suggesting that LARE is also an effective data augmentation method.

\begin{table}[ht]
\centering
\caption{Accuracy on CIFAR-100}
\label{tab:2}
\scalebox{1.0}{
\begin{tabular}{c|cc}
\toprule
Method & Accuracy [\%] & std. \\ \midrule
CoCa (zero-shot) & \multicolumn{1}{r}{74.12} & \multicolumn{1}{r}{-} \\
CoCa (fine-tuning) & \multicolumn{1}{r}{83.92} & \multicolumn{1}{r}{$\pm$0.04} \\ \midrule
LARE & \multicolumn{1}{r}{\textbf{84.03}} & \multicolumn{1}{r}{$\pm$0.04}\\
\bottomrule
\end{tabular}
}
\end{table}

\begin{figure}[ht]
\begin{center}
\includegraphics[width=\linewidth]{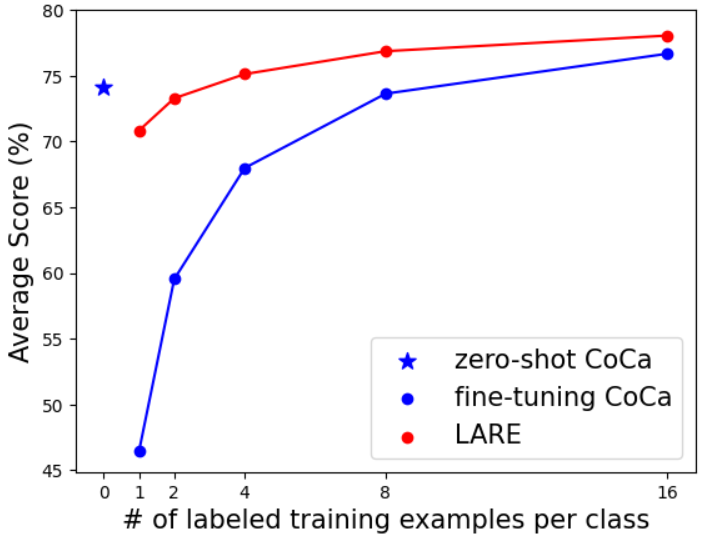}
\caption{Few-shot accuracy on CIFAR-100}
\label{fig:4}
\end{center}
\end{figure}

\paragraph*{Few-shot Learning}
Fig. \ref{fig:4} shows the few-shot accuracy on CIFAR-100 compared with CoCa to verify the effectiveness of LARE on small amounts of data. The experimental results showed that LARE outperformed CoCa (fine-tuning) in all settings, and was nearly equivalent to CoCa (fine-tuning) with four times more training data than LARE, where four originated from the sum of three augmented samples and one original data. This suggests that LARE is an image classification model that can ensure accuracy, even with small amounts of data.

\paragraph*{Results for Imbalanced Data}
Table \ref{tab:3} shows the accuracy of the imbalanced data on CIFAR-100 compared with CoCa. Imbalanced data refer to situations in which the amount of training data differs for each class, such as when it is difficult to collect images for a specific class or when there are classes that are not labeled. In this experiment, we randomly selected $X\%$ ($10\%$, $30\%$, $50\%$) of all the classes and reduced the amount of training data for these classes (originally 400) to $N$ (5, 10, and 50) to create an imbalanced dataset. We conducted experiments for $3\times3=9$ combinations of $X$ and $N$ and demonstrated the accuracy of CoCa (fine-tuning) (top) and LARE (bottom). The experimental results show that LARE outperforms CoCa (fine-tuning) in all settings by up to $1.1\%$, suggesting that LARE is an effective and versatile model for imbalanced data.

\begin{table}[ht]
\centering
\caption{Accuracy of imbalanced data on CIFAR-100. $X$ represents the percentage of classes to reduce training data to create imbalanced data, and $N$ represents the number of training data for the classes to be reduced. The top of each setting in the table shows the accuracy of CoCa (fine-tuning), and the bottom shows the accuracy of LARE. LARE can beat CoCa in all settings.}
\label{tab:3}
\begin{tabular}{c|c|c|c|c}
\toprule
\multicolumn{2}{c|}{\multirow{2}{*}{}} & \multicolumn{3}{c}{\textit{X} \%} \\ \cmidrule{3-5}
\multicolumn{2}{l|}{} & \multicolumn{1}{c|}{10 \%} & \multicolumn{1}{c|}{30 \%} & \multicolumn{1}{c}{50 \%} \\ \midrule
\multirow{6}{*}{\textit{N}} & \multirow{2}{*}{5} & \multicolumn{1}{r|}{78.22($\pm$0.32)} & \multicolumn{1}{r|}{68.51($\pm$0.83)} & \multicolumn{1}{r}{59.98($\pm$0.17)} \\
& & \multicolumn{1}{r|}{\textbf{78.49}($\pm$0.43)} & \multicolumn{1}{r|}{\textbf{69.14}($\pm$0.89)} & \multicolumn{1}{r}{\textbf{61.08}($\pm$0.33)} \\ \cmidrule{2-5}
& \multirow{2}{*}{10} & \multicolumn{1}{r|}{79.75($\pm$0.31)} & \multicolumn{1}{r|}{72.36($\pm$0.42)} & \multicolumn{1}{r}{66.68($\pm$0.57)} \\
& & \multicolumn{1}{r|}{\textbf{79.94}($\pm$0.37)} & \multicolumn{1}{r|}{\textbf{72.87}($\pm$0.44)} & \multicolumn{1}{r}{\textbf{67.31}($\pm$0.57)} \\ \cmidrule{2-5}
& \multirow{2}{*}{50} & \multicolumn{1}{r|}{82.42($\pm$0.24)} & \multicolumn{1}{r|}{79.87($\pm$0.44)} & \multicolumn{1}{r}{78.27($\pm$0.24)} \\
& & \multicolumn{1}{r|}{\textbf{82.48}($\pm$0.20)} & \multicolumn{1}{r|}{\textbf{79.99}($\pm$0.43)} & \multicolumn{1}{r}{\textbf{78.42}($\pm$0.25)} \\
\bottomrule
\end{tabular}
\end{table}

\subsection{Analysis of Latent Region}
\paragraph*{Region (Box) Size}
In this section, we analyze the region (box) created in Stage 1 of LARE. Table \ref{tab:4} lists the rankings based on the average region size for each class on CIFAR-100. The size of the region (box) is equivalent to the hypervolume of a hypercuboid and is calculated as the product of each side length. According to Table \ref{tab:4}, classes with large region sizes tend to be broad-sense and general classes, such as bear, bicycle, and train. Conversely, classes with small region sizes tend to be narrow-sense and unique classes, such as lawn-mower, skunk, and streetcar. In particular, the streetcar has a small region, whereas the train, which is a superordinate concept of the streetcar, has a large region, suggesting that multiple concepts or broad meanings can be expressed in terms of the extent of the region.

\begin{table}[ht]
\centering
\caption{Top/Bottom 10 ranking by region size. Classes are ranked by the average region size of each image on CIFAR-100.}
\label{tab:4}
\begin{tabular}{cc|cc}
\toprule
\multicolumn{2}{c|}{Top} & \multicolumn{2}{c}{Bottom}\\ \midrule
Rank & Class Name & Rank & Class Name\\ \midrule \midrule
1 & bear & 100 & lawn-mower\\
2 & turtle & 99 & sweet peppers\\
3 & motorcycle & 98 & chimpanzee\\
4 & bee & 97 & oranges\\
5 & bicycle & 96 & skunk\\
6 & spider & 95 & streetcar\\
7 & butterfly & 94 & wardrobe\\
8 & clock & 93 & cockroach\\
9 & baby & 92 & ray\\
10 & train & 91 & fox\\
\bottomrule
\end{tabular}
\end{table}

\paragraph*{Region (Box) Side Length}
Table \ref{tab:5} lists the class ranking based on the region side length for each dimension of CIFAR-100. We present three dimensions that demonstrate good characteristics. Dimension A represents animals in general, dimension B represents humans and man-made objects related to life, and dimension C represents nearby plants and objects. As each dimension has different characteristics, it can be inferred that each image is represented as a latent region with a different shape.

Based on the above, the size and side length of the regions created by LARE can be used for various downstream tasks, not just for data augmentation. For example, because the shapes of the regions are different in each image, it is conceivable to use clustering~\cite{k-means, LDA} in the same class with region size and side length as input (e.g., an image of a mouse containing flowers will have larger side lengths in dimension C as well as dimension A). This will be a subject of future research.

\begin{table}[ht]
\begin{center}
\caption{Top 5 ranking by region side length in three specific dimensions. Classes are ranked by the average region side length for each dimension on CIFAR-100.}
\label{tab:5}
\begin{tabular}{c|c|c|c}
\toprule
& Dimension A & Dimension B & Dimension C\\ \midrule
Rank & Class Name & Class Name & Class Name\\ \midrule \midrule
1 & mouse & baby & orchids\\
2 & snake & woman & road\\
3 & beetle & television & sunflowers\\
4 & elephant & tractor & tank\\
5 & turtle & house & mouse\\
\bottomrule
\end{tabular}
\end{center}
\end{table}

%---------------------------------------------------------------------
\section{Discussion}
In this section, we discuss the effectiveness of the proposed method LARE compared with LADS. In the experiment of the unseen domain on the CUB-Painting dataset, LARE's accuracy of the out-of-domain ``painting'' exceeded CoCa (fine-tuning) by up to $1.3\%$ but slightly exceeded LADS (CoCa) by only $0.7\%$ or was inferior in LADS (CLIP). This is because LADS directly generates image embeddings of a “painting” using the text prompt ``A painting of a [label].'' Conversely, because LARE augments image embeddings by randomly sampling from within the region, it is not always possible to generate image embeddings of the ``painting.'' Although LARE randomly determines the unseen domains to augment, LARE performed similar to LADS for one unseen domain of the ``painting.'' From this, it is expected that LARE will perform robust classification not only for the ``painting'' but also for various unseen domains.

Another clear difference between LARE and LADS is that LARE does not require text prompts for specific unseen domains. LADS inputs one text prompt for each unseen domain, making it difficult to apply when there are a large number of domains to consider. In fact, LADS cannot be applied to datasets, such as CIFAR-100, which do not have specific unseen domains; in particular, they have numerous unseen domains to consider. However, our proposed method LARE can augment data to various unseen domains without a text prompt for specific unseen domains, making it a versatile model that can be used in various situations.

%---------------------------------------------------------------------
%\section{Ablation}
%\subsection{n}

%\subsection{a}

%---------------------------------------------------------------------
\section{Conclusion and Limitation}
In this study, we present LARE as a novel and robust image classification model that applies regional embedding to a VLM. LARE augments data from within the latent region to various domains by utilizing the richness of the embedding space trained in the pre-trained VLM, adapting to unseen domains and improving the accuracy compared with previous fine-tuning models. In addition, experiments conducted under multiple conditions, such as small amounts of data, imbalanced data, and region shape analysis, suggest that LARE is a versatile image classification model. 

A limitation of LARE is that it relies on the richness of the VLM embedding space. LARE cannot be expected to achieve significantly better accuracy than the previous models. However, as larger or more accurate VLMs are developed, our model will improve accuracy along with them and our study's results are highly valuable in such prospects. In future work, we expect to augment more reliable embedding by improving the method of creating regions or losses. Furthermore, we hope that LARE will develop into a more effective and innovative method by leveraging LARE's strengths of extensive and persistent data augmentation.

\section*{Acknowledgment}
This work was supported by JSPS, Japan KAKENHI Grant Numbers 21H04600 and 24H00370.

\section*{Declaration of competing interest}
The authors declare that they have no known competing financial interests or personal relationships that could have appeared to influence the work reported in this paper.
%--------------------------------------------------------------------------------------------------------------------------------

%%Bibliography
\bibliographystyle{unsrt}  
\bibliography{references}  

\begin{thebibliography}{10}

\bibitem{CUB}
Catherine Wah, Steve Branson, Peter Welinder, Pietro Perona, and Serge
  Belongie.
\newblock The caltech-ucsd birds-200-2011 dataset.
\newblock 2011.

\bibitem{CUB-Painting}
Sinan Wang, Xinyang Chen, Yunbo Wang, Mingsheng Long, and Jianmin Wang.
\newblock Progressive adversarial networks for fine-grained domain adaptation.
\newblock In {\em Proceedings of CVPR}, pages 9213--9222, 2020.

\bibitem{CLIP}
Alec Radford, Jong~Wook Kim, Chris Hallacy, Aditya Ramesh, Gabriel Goh,
  Sandhini Agarwal, Girish Sastry, Amanda Askell, Pamela Mishkin, Jack Clark,
  et~al.
\newblock Learning transferable visual models from natural language
  supervision.
\newblock In {\em Proceedings of ICML}, pages 8748--8763, 2021.

\bibitem{CoCa}
Jiahui Yu, Zirui Wang, Vijay Vasudevan, Legg Yeung, Mojtaba Seyedhosseini, and
  Yonghui Wu.
\newblock Coca: Contrastive captioners are image-text foundation models.
\newblock In {\em Proceedings of CVPR}, 2022.

\bibitem{ALIGN}
Chao Jia, Yinfei Yang, Ye~Xia, Yi-Ting Chen, Zarana Parekh, Hieu Pham, Quoc Le,
  Yun-Hsuan Sung, Zhen Li, and Tom Duerig.
\newblock Scaling up visual and vision-language representation learning with
  noisy text supervision.
\newblock In {\em Proceedings of ICML}, pages 4904--4916. PMLR, 2021.

\bibitem{FILIP}
Lewei Yao, Runhui Huang, Lu~Hou, Guansong Lu, Minzhe Niu, Hang Xu, Xiaodan
  Liang, Zhenguo Li, Xin Jiang, and Chunjing Xu.
\newblock Filip: Fine-grained interactive language-image pre-training.
\newblock In {\em Proceedings of ICLR}, 2022.

\bibitem{Flamingo}
Jean-Baptiste Alayrac, Jeff Donahue, Pauline Luc, Antoine Miech, Iain Barr,
  Yana Hasson, Karel Lenc, Arthur Mensch, Katherine Millican, Malcolm Reynolds,
  et~al.
\newblock Flamingo: a visual language model for few-shot learning.
\newblock {\em Advances in NeurIPS}, 35:23716--23736, 2022.

\bibitem{BLIP}
Junnan Li, Dongxu Li, Caiming Xiong, and Steven Hoi.
\newblock Blip: Bootstrapping language-image pre-training for unified
  vision-language understanding and generation.
\newblock In {\em Proceedings of ICML}, pages 12888--12900. PMLR, 2022.

\bibitem{BLIP2}
Junnan Li, Dongxu Li, Silvio Savarese, and Steven Hoi.
\newblock Blip-2: Bootstrapping language-image pre-training with frozen image
  encoders and large language models.
\newblock In {\em Proceedings of ICML}, pages 19730--19742. PMLR, 2023.

\bibitem{CoOp}
Kaiyang Zhou, Jingkang Yang, Chen~Change Loy, and Ziwei Liu.
\newblock Learning to prompt for vision-language models.
\newblock {\em Advances in IJCV}, 130(9):2337--2348, 2022.

\bibitem{Clip-adapter}
Peng Gao, Shijie Geng, Renrui Zhang, Teli Ma, Rongyao Fang, Yongfeng Zhang,
  Hongsheng Li, and Yu~Qiao.
\newblock Clip-adapter: Better vision-language models with feature adapters.
\newblock {\em Advances in IJCV}, 132(2):581--595, 2024.

\bibitem{Tip-adapter}
Renrui Zhang, Rongyao Fang, Wei Zhang, Peng Gao, Kunchang Li, Jifeng Dai,
  Yu~Qiao, and Hongsheng Li.
\newblock Tip-adapter: Training-free clip-adapter for better vision-language
  modeling.
\newblock In {\em Proceedings of ECCV}, 2022.

\bibitem{UPL}
Tony Huang, Jack Chu, and Fangyun Wei.
\newblock Unsupervised prompt learning for vision-language models.
\newblock {\em arXiv preprint arXiv:2204.03649}, 2022.

\bibitem{Hazardous}
Ran Zhang, Zhila Bahrami, Ke~Feng, and Zheng Liu.
\newblock A visual and textual information fusion-based zero-shot framework for
  hazardous material placard detection and recognition.
\newblock {\em IEEE Transactions on Artificial Intelligence}, 2023.

\bibitem{Fine-tuning}
Ananya Kumar, Aditi Raghunathan, Robbie Jones, Tengyu Ma, and Percy Liang.
\newblock Fine-tuning can distort pretrained features and underperform
  out-of-distribution.
\newblock In {\em Proceedings of ICLR}, 2022.

\bibitem{FAKE}
Mert~Bulent Sariyildiz, Karteek Alahari, Diane Larlus, and Yannis Kalantidis.
\newblock Fake it till you make it: Learning transferable representations from
  synthetic imagenet clones.
\newblock In {\em Proceedings of CVPR}, 2023.

\bibitem{ALIA}
Lisa Dunlap, Alyssa Umino, Han Zhang, Jiezhi Yang, Joseph~E Gonzalez, and
  Trevor Darrell.
\newblock Diversify your vision datasets with automatic diffusion-based
  augmentation.
\newblock In {\em Proceedings of CVPR}, 2023.

\bibitem{Attributed}
Shijian Wang, Linxin Song, Ryotaro Shimizu, Masayuki Goto, and Hanqian wu.
\newblock Attributed synthetic data generation for zero-shot image
  classification.
\newblock In {\em Proceedings of CVPR}, 2024.

\bibitem{StableDiffusion}
Robin Rombach, Andreas Blattmann, Dominik Lorenz, Patrick Esser, and Bj{\"o}rn
  Ommer.
\newblock High-resolution image synthesis with latent diffusion models.
\newblock In {\em Proceedings of CVPR}, pages 10684--10695, 2022.

\bibitem{DALL-E}
Aditya Ramesh, Mikhail Pavlov, Gabriel Goh, Scott Gray, Chelsea Voss, Alec
  Radford, Mark Chen, and Ilya Sutskever.
\newblock Zero-shot text-to-image generation.
\newblock In {\em Proceedings of ICML}, pages 8821--8831. Pmlr, 2021.

\bibitem{DALL-E2}
Aditya Ramesh, Prafulla Dhariwal, Alex Nichol, Casey Chu, and Mark Chen.
\newblock Hierarchical text-conditional image generation with clip latents.
\newblock {\em arXiv preprint arXiv:2204.06125}, 1(2):3, 2022.

\bibitem{DALL-E3}
James Betker, Gabriel Goh, Li~Jing, TimBrooks, Jianfeng Wang, Linjie Li,
  LongOuyang, JuntangZhuang, JoyceLee, YufeiGuo, WesamManassra,
  PrafullaDhariwal, CaseyChu, YunxinJiao, and Aditya Ramesh.
\newblock Improving image generation with better captions.
\newblock 2023.

\bibitem{liu2024best}
Ruibo Liu, Jerry Wei, Fangyu Liu, Chenglei Si, Yanzhe Zhang, Jinmeng Rao,
  Steven Zheng, Daiyi Peng, Diyi Yang, Denny Zhou, et~al.
\newblock Best practices and lessons learned on synthetic data for language
  models.
\newblock {\em arXiv preprint arXiv:2404.07503}, 2024.

\bibitem{LADS}
Lisa Dunlap, Clara Mohri, Devin Guillory, Han Zhang, Trevor Darrell, Joseph~E
  Gonzalez, Aditi Raghunathan, and Anja Rohrbach.
\newblock Using language to extend to unseen domains.
\newblock In {\em Proceedings of ICLR}, 2023.

\bibitem{TextManiA}
Moon Ye-Bin, Jisoo Kim, Hongyeob Kim, Kilho Son, and Tae-Hyun Oh.
\newblock Textmania: Enriching visual feature by text-driven manifold
  augmentation.
\newblock In {\em \textit{Proceedings of ICCV}}, pages 2526--2537, 2023.

\bibitem{Poda}
Mohammad Fahes, Tuan-Hung Vu, Andrei Bursuc, Patrick P{\'e}rez, and Raoul
  De~Charette.
\newblock Poda: Prompt-driven zero-shot domain adaptation.
\newblock In {\em Proceedings of ICCV}, pages 18623--18633, 2023.

\bibitem{LanDA}
Zhenbin Wang, Lei Zhang, Lituan Wang, and Minjuan Zhu.
\newblock Landa: Language-guided multi-source domain adaptation.
\newblock {\em arXiv preprint arXiv:2401.14148}, 2024.

\bibitem{Word2Box}
Shib~Sankar Dasgupta, Michael Boratko, Siddhartha Mishra, Shriya Atmakuri,
  Dhruvesh Patel, Xiang~Lorraine Li, and Andrew McCallum.
\newblock Word2box: Capturing set-theoretic semantics of words using box
  embeddings.
\newblock In {\em Proceedings of ACL}, pages 2263--2276, 2022.

\bibitem{MBM}
Dhruvesh Patel, Pavitra Dangati, Jay-Yoon Lee, Michael Boratko, and Andrew
  McCallum.
\newblock Modeling label space interactions in multi-label classification using
  box embeddings.
\newblock In {\em Proceedings of ICLR}, 2022.

\bibitem{BoxTaxo}
Song Jiang, Qiyue Yao, Qifan Wang, and Yizhou Sun.
\newblock A single vector is not enough: Taxonomy expansion via box embeddings.
\newblock In {\em Proceedings of ACM}, pages 2467--2476, 2023.

\bibitem{DomainNet}
Xingchao Peng, Qinxun Bai, Xide Xia, Zijun Huang, Kate Saenko, and Bo~Wang.
\newblock Moment matching for multi-source domain adaptation.
\newblock In {\em Proceedings of ICCV}, pages 1406--1415, 2019.

\bibitem{CIFAR-100}
Alex Krizhevsky, Geoffrey Hinton, et~al.
\newblock Learning multiple layers of features from tiny images.
\newblock 2009.

\bibitem{contrastive}
Ashish Jaiswal, Ashwin~Ramesh Babu, Mohammad~Zaki Zadeh, Debapriya Banerjee,
  and Fillia Makedon.
\newblock A survey on contrastive self-supervised learning.
\newblock {\em Technologies}, 9(1), 2021.

\bibitem{Vilbert}
Jiasen Lu, Dhruv Batra, Devi Parikh, and Stefan Lee.
\newblock Vilbert: Pretraining task-agnostic visiolinguistic representations
  for vision-and-language tasks.
\newblock {\em Advances in NeurIPS}, 32, 2019.

\bibitem{Transformer}
Ashish Vaswani, Noam Shazeer, Niki Parmar, Jakob Uszkoreit, Llion Jones,
  Aidan~N Gomez, {\L}ukasz Kaiser, and Illia Polosukhin.
\newblock Attention is all you need.
\newblock {\em Advances in NeurIPS}, 30, 2017.

\bibitem{Vit}
Alexey Dosovitskiy, Lucas Beyer, Alexander Kolesnikov, Dirk Weissenborn,
  Xiaohua Zhai, Thomas Unterthiner, Mostafa Dehghani, Matthias Minderer, Georg
  Heigold, Sylvain Gelly, et~al.
\newblock An image is worth 16x16 words: Transformers for image recognition at
  scale.
\newblock In {\em Proceedings of ICLR}, 2021.

\bibitem{SimVLM}
Zirui Wang, Jiahui Yu, Adams~Wei Yu, Zihang Dai, Yulia Tsvetkov, and Yuan Cao.
\newblock Simvlm: Simple visual language model pretraining with weak
  supervision.
\newblock In {\em Proceedings of ICLR}, 2022.

\bibitem{DA1}
Youngeun Kim, Donghyeon Cho, Kyeongtak Han, Priyadarshini Panda, and Sungeun
  Hong.
\newblock Domain adaptation without source data.
\newblock {\em IEEE Transactions on Artificial Intelligence}, 2(6):508--518,
  2021.

\bibitem{domain-adaptation1}
Benjamin Recht, Rebecca Roelofs, Ludwig Schmidt, and Vaishaal Shankar.
\newblock Do imagenet classifiers generalize to imagenet?
\newblock In {\em Proceedings of ICML}, pages 5389--5400. PMLR, 2019.

\bibitem{domain-adaptation2}
Zak Murez, Soheil Kolouri, David Kriegman, Ravi Ramamoorthi, and Kyungnam Kim.
\newblock Image to image translation for domain adaptation.
\newblock In {\em Proceedings of CVPR}, pages 4500--4509, 2018.

\bibitem{Visda}
Xingchao Peng, Ben Usman, Neela Kaushik, Judy Hoffman, Dequan Wang, and Kate
  Saenko.
\newblock Visda: The visual domain adaptation challenge.
\newblock {\em arXiv preprint arXiv:1710.06924}, 2017.

\bibitem{domain-adaptation3}
Yaroslav Ganin and Victor Lempitsky.
\newblock Unsupervised domain adaptation by backpropagation.
\newblock In {\em Proceedings of ICML}, pages 1180--1189. PMLR, 2015.

\bibitem{ADDA}
Eric Tzeng, Judy Hoffman, Kate Saenko, and Trevor Darrell.
\newblock Adversarial discriminative domain adaptation.
\newblock In {\em Proceedings of CVPR}, pages 7167--7176, 2017.

\bibitem{domain-adaptation4}
Emadeldeen Eldele, Mohamed Ragab, Zhenghua Chen, Min Wu, Chee-Keong Kwoh, and
  Xiaoli Li.
\newblock Contrastive domain adaptation for time-series via temporal mixup.
\newblock {\em IEEE Transactions on Artificial Intelligence}, 2023.

\bibitem{AD-clip}
Mainak Singha, Harsh Pal, Ankit Jha, and Biplab Banerjee.
\newblock Ad-clip: Adapting domains in prompt space using clip.
\newblock In {\em Proceedings of ICCV}, pages 4355--4364, 2023.

\bibitem{DPL}
Xin Zhang, Shixiang~Shane Gu, Yutaka Matsuo, and Yusuke Iwasawa.
\newblock Domain prompt learning for efficiently adapting clip to unseen
  domains.
\newblock {\em Transactions of JSAI}, 38(6), 2023.

\bibitem{WiSE-FT}
Mitchell Wortsman, Gabriel Ilharco, Jong~Wook Kim, Mike Li, Simon Kornblith,
  Rebecca Roelofs, Raphael~Gontijo Lopes, Hannaneh Hajishirzi, Ali Farhadi,
  Hongseok Namkoong, et~al.
\newblock Robust fine-tuning of zero-shot models.
\newblock In {\em Proceedings of CVPR}, pages 7959--7971, 2022.

\bibitem{Padclip}
Zhengfeng Lai, Noranart Vesdapunt, Ning Zhou, Jun Wu, Cong~Phuoc Huynh, Xuelu
  Li, Kah~Kuen Fu, and Chen-Nee Chuah.
\newblock Padclip: Pseudo-labeling with adaptive debiasing in clip for
  unsupervised domain adaptation.
\newblock In {\em Proceedings of ICCV}, pages 16155--16165, 2023.

\bibitem{DAPrompt}
Chunjiang Ge, Rui Huang, Mixue Xie, Zihang Lai, Shiji Song, Shuang Li, and Gao
  Huang.
\newblock Domain adaptation via prompt learning.
\newblock {\em IEEE Transactions on NNLS}, 2023.

\bibitem{ReCLIP}
Xuefeng Hu, Ke~Zhang, Lu~Xia, Albert Chen, Jiajia Luo, Yuyin Sun, Ken Wang, Nan
  Qiao, Xiao Zeng, Min Sun, et~al.
\newblock Reclip: Refine contrastive language image pre-training with source
  free domain adaptation.
\newblock In {\em Proceedings of ACV}, pages 2994--3003, 2024.

\bibitem{Promptstyler}
Junhyeong Cho, Gilhyun Nam, Sungyeon Kim, Hunmin Yang, and Suha Kwak.
\newblock Promptstyler: Prompt-driven style generation for source-free domain
  generalization.
\newblock In {\em Proceedings of ICCV}, pages 15702--15712, 2023.

\bibitem{Clipood}
Yang Shu, Xingzhuo Guo, Jialong Wu, Ximei Wang, Jianmin Wang, and Mingsheng
  Long.
\newblock Clipood: Generalizing clip to out-of-distributions.
\newblock In {\em Proceedings of ICML}, pages 31716--31731. PMLR, 2023.

\bibitem{Styleclip}
Or~Patashnik, Zongze Wu, Eli Shechtman, Daniel Cohen-Or, and Dani Lischinski.
\newblock Styleclip: Text-driven manipulation of stylegan imagery.
\newblock In {\em Proceedings of ICCV}, pages 2085--2094, 2021.

\bibitem{Stylegan-nada}
Rinon Gal, Or~Patashnik, Haggai Maron, Amit~H Bermano, Gal Chechik, and Daniel
  Cohen-Or.
\newblock Stylegan-nada: Clip-guided domain adaptation of image generators.
\newblock {\em Transactions on ACM}, 41(4):1--13, 2022.

\bibitem{COAL}
Shuhan Tan, Xingchao Peng, and Kate Saenko.
\newblock Class-imbalanced domain adaptation: An empirical odyssey.
\newblock In {\em Proceedings of ECCV}, pages 585--602. Springer, 2020.

\bibitem{AdamW}
Ilya Loshchilov and Frank Hutter.
\newblock Decoupled weight decay regularization.
\newblock In {\em Proceedings of ICLR}, 2019.

\bibitem{k-means}
Stuart Lloyd.
\newblock Least squares quantization in pcm.
\newblock {\em IEEE transactions on information theory}, 28(2):129--137, 1982.

\bibitem{LDA}
David~M Blei, Andrew~Y Ng, and Michael~I Jordan.
\newblock Latent dirichlet allocation.
\newblock {\em Journal of machine Learning research}, 3(Jan):993--1022, 2003.

\end{thebibliography}

%\input{references.bbl}

%--------------------------------------------------------------------------------------------------------------------------------
%\section*{Appendix}

\end{document}